\pdfoutput=1

\documentclass[11pt]{article}

\usepackage[final]{ACL2023}

\usepackage{times}
\usepackage{latexsym}
\usepackage{graphics}
\usepackage{graphicx}
\usepackage[T1]{fontenc}
\usepackage[utf8]{inputenc}

\usepackage[T1]{fontenc}

\usepackage{microtype}

\usepackage{inconsolata}

\title{Gender-tuning: Empowering Fine-tuning for Debiasing Pre-trained Language Models}

\author{Somayeh Ghanbarzadeh \\
  University of North Texas\\
  \texttt{somayehghanbarzadeh@my.unt.edu} \\\And
   Yan Huang \\
 University of North Texas \\
  \texttt{yan.huangl@unt.edu} \\ \And
 Hamid Palangi\\
  Microsoft Research\\
  \texttt{hpalangi@microsoft.com} \\\AND
   Radames Cruz Moreno\\
  Microsoft Research\\
  \texttt{radames.cruz@microsoft.com} \\\And
  Hamed Khanpour\\
  Microsoft Research\\
  \texttt{hamed.khanpour@microsoft.com} \\
  }
\begin{document}
\maketitle
\begin{abstract}
Recent studies have revealed that the widely-used Pre-trained Language Models (PLMs) propagate societal biases from the large unmoderated pre-training corpora. Existing solutions require debiasing training processes and datasets for debiasing, which are resource-intensive and costly. Furthermore,  these methods hurt the PLMs' performance on downstream tasks. In this study, we propose \emph{Gender-tuning}, which debiases the PLMs through fine-tuning on downstream tasks' datasets. For this aim, Gender-tuning integrates Masked Language Modeling (MLM) training objectives into fine-tuning's training process. Comprehensive experiments show that Gender-tuning outperforms the state-of-the-art baselines in terms of average gender bias scores in PLMs while improving PLMs' performance on downstream tasks solely using the downstream tasks' dataset. Also, Gender-tuning is a deployable debiasing tool for any PLM that works with original fine-tuning.
\end{abstract}

\section{Introduction}
Pre-trained Language Models (PLMs) have achieved state-of-the-art performance across various tasks in natural language processing \citep{devlin2019bert, liu2019roberta, clark2020electra}. One of the crucial reasons for this success is pre-training on large-scale corpora, which is collected from unmoderated sources such as the internet. Prior studies \citep{caliskan2017semantics, zhao2018learning, may2019measuring, kurita2019measuring, gehman2020realtoxicityprompts}  have shown that PLMs capture a significant amount of social biases existing in the pre-training corpus. For instance, they showed that the PLMs learned that the word "he" is closer to the word "engineer" because of the frequent co-occurrence of this combination in the training corpora, which is known as social gender biases. Since PLMs are increasingly deployed in real-world scenarios, there is a serious concern that they propagate discriminative prediction and unfairness.

Several solutions for mitigating the social biases have been proposed, including: using banned word lists \citep{raffel2020exploring}, building deliberated training datasets \citep{bender2021dangers}, balancing the biased and unbiased terms in the training dataset \citep{dixon2018measuring,bordia2019identifying}, debiasing embedding spaces \citep{liang2020towards, cheng2021fairfil}, and self-debiasing in text generation \citep{schick2021self}. Although all these solutions have shown different levels of success, they tend to limit the PLMs' ability \citep{meade2022empirical}. For example, the banned words solution prevent gaining knowledge of topics related to banned words. Also,  some of them hurt the PLMs' performance on downstream tasks. Furthermore, dataset curation and pre-training are two resource-intensive tasks needed for most of the above solutions \citep{schick2021self}.

In this study, we address the challenges mentioned above by proposing an effective approach named \emph{Gender-tuning} for debiasing the PLMs through fine-tuning on downstream tasks' datasets. For this goal, Gender-tuning perturbs the training examples by first finding the gender-words in the training examples based on a given gender-word list. Then Gender-tuning replaces them with the new words to interrupt the association between the gender-words and other words in the training examples (Table \ref{tab:table4}). Finally, Gender-tuning classifies the examples with the replaced words according to the original training examples' ground-truth labels to compute a joint loss from perturbation and classification for training the Gender-tuning.    
\begin{figure*}[t]
\centering
\includegraphics[width=0.7\linewidth]{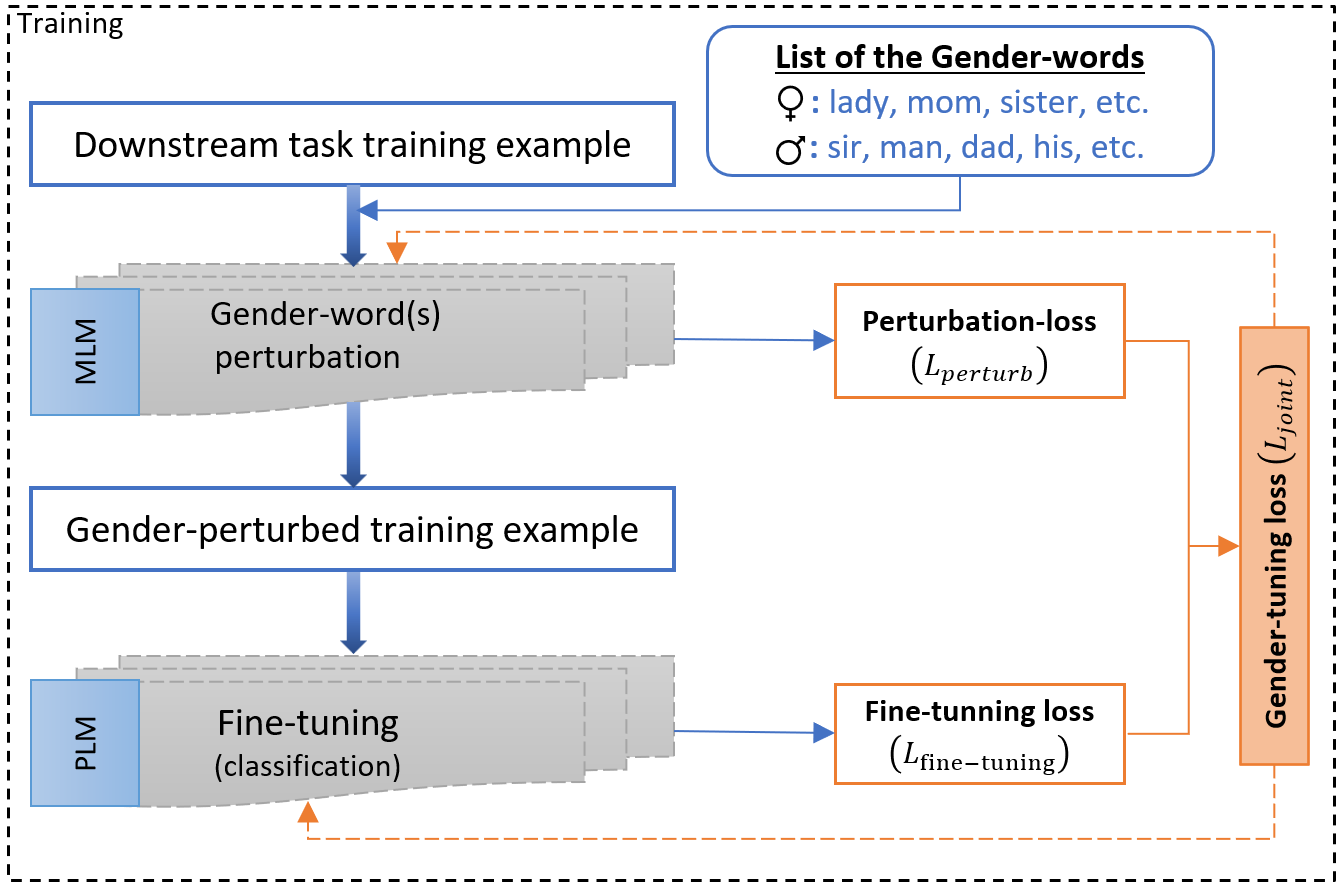}
   
\caption{ Illustration of Gender-tuning training process. MLM and PLM be trained based on Gender-tuning loss. The examples without any gender-word are directly fed to fine-tuning.  }
\label{fig:fig2}
\end{figure*}

The key advantage of our method is integrating the debiasing process into the fine-tuning that allows the debiasing and fine-tuning to perform simultaneously. Thus, Gender-tuning does not require separate pre-training or additional training data. Also, this integration makes Gender-tuning a plug-and-play debiasing tool for any PLMs that works with original fine-tuning.
\begin{table*}[t]
\centering
\begin{tabular}{l l}
\hline\hline
 &\multicolumn{1}{c}{}\\
 1&\multicolumn{1}{l }{\textbf{Original example:}} \\
 &\multicolumn{1}{l}{"[\textbf{he}] is at 22 a powerful [\textbf{actor}]."}\\
 &\multicolumn{1}{l }{\textbf{Perturbed examples:}}\\
 &\multicolumn{1}{l }{  epoch 1 $\Rightarrow$  "[\textbf{\color{blue}{girl}}] is at 22 a powerful [\textbf{\color{blue}{UNK}}]."}\\
 &\multicolumn{1}{l }{ epoch 2 $\Rightarrow$  "[\textbf{\color{blue}{boy}}] is at 22 a powerful [\textbf{\color{blue}{actor}}]."}\\
 &\multicolumn{1}{l }{  epoch 3 $\Rightarrow$  "[\textbf{\color{blue}{She}}] is at 22 a powerful [\textbf{\color{blue}{actress}}]."}\\
 &\multicolumn{1}{c}{}\\
 2&\multicolumn{1}{l}{\textbf{Original example:}}\\
 &\multicolumn{1}{l }{ "[\textbf{she}] beautifully chaperon the [\textbf{girls}] in the kitchen."} \\
 &\multicolumn{1}{l}{\textbf{Perturbed examples:}}\\
 &\multicolumn{1}{l }{    epoch 1 $\Rightarrow$   "[\textbf{l\color{blue}{ady}}] beautifully chaperon the [\textbf{\color{blue}{women}}] in the kitchen."}\\
 &\multicolumn{1}{l }{  epoch 2 $\Rightarrow$  "[\textbf{\color{blue}{girl}}] beautifully chaperon the [\textbf{\color{blue}{boys}}] in the kitchen."}\\
 &\multicolumn{1}{l }{  epoch 3 $\Rightarrow$   "[\textbf{\color{blue}{he}}] beautifully chaperon the [\textbf{\color{blue}{men}}] in the kitchen."}\\
 &\multicolumn{1}{l}{}\\ \hline \hline
 
\end{tabular}
\caption{Some perturbed examples generated by Gender-tuning through three training epochs.}
\label{tab:table4}
\end{table*} 



To evaluate the effectiveness of our proposed method, we conducted comprehensive experiments following two state-of-the-art debiasing baselines: SENT-DEBIAS (Sent-D) \citep{liang2020towards} and FairFil (FairF) \citep{cheng2021fairfil}. The results show that Gender-tuning outperforms both baselines in terms of the average gender-bias scores in the BERT model while improving its performance on the downstream tasks. In addition, we reported the performance of Gender-tuning applied to the RoBERTa that shows considerable improvement.
Finally, our ablation studies demonstrate that all components of  Gender-tuning, including two training phases and joint loss, play an essential role in achieving success. 
\section{Methodology}
We propose a novel debiasing approach, named Gender-tuning (Figure \ref{fig:fig2}), that performs the debiasing process and fine-tuning simultaneously on the downstream tasks' dataset.
For this aim, Gender-tuning integrates two training objectives: 1) Masked Language Modeling (MLM) training objective for gender-word perturbation and 2) Fine-tuning for classification. In each training batch, Gender-tuning works as follows:\\
Gender-tuning uses MLM to perturb training examples by masking the existing gender-word(s). For gender-words, we use the feminine and masculine word lists created by \cite{zhao2018learning}. The MLM training objective is to predict masked token(s) with a mean cross-entropy loss  that we denote as perturbation-loss ($\mathcal{L}_{perturb}$). The training examples with predicted tokens, called \emph{gender-perturbed examples} (Table \ref{tab:table4}),  are fed into fine-tuning to be classified according to the original examples' ground-truth label ($y$).
Then $p_{\theta}(y' = y | \hat{x})$ is the fine-tuning classification function to predict the gender-perturbed example's label ($y'$) based on the gender-perturbed example ($\hat{x}$) to compute the fine-tuning loss ($\mathcal{L}_{fine-tuning}$), where $\theta$ is the PLM's parameters for the fine-tuning. A weighted aggregation of the perturbation loss and fine-tuning loss, called joint-loss ($\mathcal{L}_{joint}$), is used for training the Gender-tuning as follows:
\begin{equation}
\mathcal{L}_{joint} = \alpha \  \mathcal{L}_{perturb} + \:(1-\alpha) \mathcal{L}_{fine-tuning}
\label{eq:1}
\end{equation} 
where ${\alpha}$ is a weighting factor that is employed to adjust the contribution of the two training losses in computing the joint-loss. 

The Gender-tuning training objective is to minimize joint-loss to ensure that the label of the perturbed example is the same as the label of the original training example. In the following, we present how joint-loss impacts the training process of Gender-tuning in each training batch:



\begin{table*}[t]
\centering
\resizebox{13.8cm}{!}{
\begin{tabular}{l|c| c c c c ccc}
\hline

\multicolumn{1}{l}{\textbf{SST-2}} & \multicolumn{5}{c}{\textbf{BERT }} & \multicolumn{3}{c}{\textbf{RoBERTa }} \\ \hline
\multicolumn{1}{l|}{}&\multicolumn{1}{l|}{Origin} & \multicolumn{1}{c}{Sent-D}&  \multicolumn{1}{c}{FairF}& \multicolumn{1}{l}{Gender-tuning$_{random}$}&\multicolumn{1}{l|}{Gender-tuning (ours)}&  \multicolumn{1}{l|}{Origin} & \multicolumn{1}{l}{Gender-tuning$_{random}$}&\multicolumn{1}{l}{Gender-tuning (ours)} \\ \hline
\multicolumn{1}{l|}{Names, Career/Family} & \multicolumn{1}{c|}{0.03} & \multicolumn{1}{c}{0.10}& \multicolumn{1}{c}{0.21}& \multicolumn{1}{c}{0.46}&\multicolumn{1}{c|}{\textbf{0.03}}& \multicolumn{1}{c|}{0.07} & \multicolumn{1}{c}{\textbf{0.08}}&\multicolumn{1}{c}{0.14}\\
\multicolumn{1}{l|}{Terms, Career/Family} & \multicolumn{1}{c|}{0.01} &\multicolumn{1}{c}{0.05}&\multicolumn{1}{c}{0.37} &\multicolumn{1}{c}{\textbf{0.03}}&\multicolumn{1}{c|}{0.16}& \multicolumn{1}{c|}{0.33} & \multicolumn{1}{c}{0.44}&\multicolumn{1}{c}{\textbf{0.01}}\\
\multicolumn{1}{l|}{Terms, Math/Art} &\multicolumn{1}{c|}{0.21} &\multicolumn{1}{c}{0.22}&\multicolumn{1}{c}{0.26}& \multicolumn{1}{c}{\textbf{0.05}}&\multicolumn{1}{c|}{0.39}&  \multicolumn{1}{c|}{1.32} & \multicolumn{1}{c}{1.25}&\multicolumn{1}{c}{\textbf{0.57}}\\
\multicolumn{1}{l|}{Names, Math/Art} &\multicolumn{1}{c|}{1.15} &\multicolumn{1}{c}{0.75}&\multicolumn{1}{c}{\textbf{0.09}}& \multicolumn{1}{c}{0.65}&\multicolumn{1}{c|}{0.31}&  \multicolumn{1}{c|}{1.34} & \multicolumn{1}{c}{1.12}&\multicolumn{1}{c}{\textbf{1.11}}\\
\multicolumn{1}{l|}{Terms, Science/Art} &\multicolumn{1}{c|}{0.10} &\multicolumn{1}{c}{0.08}&\multicolumn{1}{c}{0.12}& \multicolumn{1}{c}{0.42}&\multicolumn{1}{c|}{\textbf{0.07}} & \multicolumn{1}{c|}{0.25} & \multicolumn{1}{c}{\textbf{0.12}}&\multicolumn{1}{c}{0.47}\\
\multicolumn{1}{l|}{Names, Science/Art} &\multicolumn{1}{c|}{0.22} &\multicolumn{1}{c}{0.04}&\multicolumn{1}{c}{\textbf{0.005}}& \multicolumn{1}{c}{0.38}&\multicolumn{1}{c|}{0.10}&  \multicolumn{1}{c|}{0.47} & \multicolumn{1}{c}{0.62}&\multicolumn{1}{c}{\textbf{0.47}}\\\hline
\multicolumn{1}{l|}{Avg. Abs. e-size} &\multicolumn{1}{c|}{0.291} &\multicolumn{1}{c}{0.212}&\multicolumn{1}{c}{0.182}& \multicolumn{1}{c}{0.331}&\multicolumn{1}{c|}{\textbf{0.176}}&  \multicolumn{1}{c|}{0.630} & \multicolumn{1}{c}{0.605}&\multicolumn{1}{c}{\textbf{0.461}}\\\hline
\multicolumn{1}{l|}{Accuracy} &\multicolumn{1}{c|}{91.97} &\multicolumn{1}{c}{89.10}&\multicolumn{1}{c}{91.60}& \multicolumn{1}{c}{\textbf{92.66}}&\multicolumn{1}{c|}{92.10}&  \multicolumn{1}{c|}{93.57 } & \multicolumn{1}{c}{\textbf{93.92}}&\multicolumn{1}{c}{93.69}\\\hline\hline
\multicolumn{1}{l}{\textbf{CoLA}} & \multicolumn{5}{c}{\textbf{ }}   & \multicolumn{3}{c}{\textbf{ }}\\ \hline
\multicolumn{1}{l|}{Names, Career/Family} &\multicolumn{1}{c|}{0.009} &\multicolumn{1}{c}{0.14}&\multicolumn{1}{c}{\textbf{0.03}}& \multicolumn{1}{c}{0.34}&\multicolumn{1}{c|}{0.09}& \multicolumn{1}{c|}{0.29} & \multicolumn{1}{c}{0.15}&\multicolumn{1}{c}{\textbf{0.05}}\\
\multicolumn{1}{l|}{Terms, Career/Family} &\multicolumn{1}{c|}{0.19} &\multicolumn{1}{c}{0.18}&\multicolumn{1}{c}{0.11}& \multicolumn{1}{c}{0.15}&\multicolumn{1}{c|}{\textbf{0.03}} &\multicolumn{1}{c|}{0.26} & \multicolumn{1}{c}{0.08}&\multicolumn{1}{c}{\textbf{0.00}}\\
\multicolumn{1}{l|}{Terms, Math/Art} &\multicolumn{1}{c|}{0.26}  &\multicolumn{1}{c}{0.31}&\multicolumn{1}{c}{0.09}& \multicolumn{1}{c}{0.55}&\multicolumn{1}{c|}{\textbf{0.08}}& \multicolumn{1}{c|}{0.06} & \multicolumn{1}{c}{\textbf{0.02}}&\multicolumn{1}{c}{0.15}\\
\multicolumn{1}{l|}{Names, Math/Art}  &\multicolumn{1}{c|}{0.15} &\multicolumn{1}{c}{0.30}&\multicolumn{1}{c}{\textbf{0.10}}& \multicolumn{1}{c}{0.72}&\multicolumn{1}{c|}{0.24}& \multicolumn{1}{c|}{0.06} & \multicolumn{1}{c}{0.25}&\multicolumn{1}{c}{\textbf{0.07}}\\
\multicolumn{1}{l|}{Terms, Science/Art} & \multicolumn{1}{c|}{0.42} &\multicolumn{1}{c}{0.16}&\multicolumn{1}{c}{0.24}& \multicolumn{1}{c}{\textbf{0.05}}&\multicolumn{1}{c|}{0.07}& \multicolumn{1}{c|}{0.32} & \multicolumn{1}{c}{\textbf{0.57}}&\multicolumn{1}{c}{0.70}\\
\multicolumn{1}{l|}{Names, Science/Art}  &\multicolumn{1}{c|}{0.03} &\multicolumn{1}{c}{0.19}&\multicolumn{1}{c}{0.12}& \multicolumn{1}{c}{0.28}&\multicolumn{1}{c|}{\textbf{0.07}}& \multicolumn{1}{c|}{0.27} & \multicolumn{1}{c}{0.14}&\multicolumn{1}{c}{\textbf{0.03}}\\\hline
\multicolumn{1}{l|}{Avg. Abs. e-size} &\multicolumn{1}{c|}{0.181} &\multicolumn{1}{c}{.217}&\multicolumn{1}{c}{0.120}& \multicolumn{1}{c}{0.343}&\multicolumn{1}{c|}{\textbf{0.096}}&  \multicolumn{1}{c|}{0.210} & \multicolumn{1}{c}{0.201}&\multicolumn{1}{c}{\textbf{0.166}}\\\hline
\multicolumn{1}{l|}{Accuracy} &\multicolumn{1}{c|}{56.51} &\multicolumn{1}{c}{55.40}&\multicolumn{1}{c}{56.50}& \multicolumn{1}{c}{\textbf{56.85}}&\multicolumn{1}{c|}{56.60}&  \multicolumn{1}{c|}{57.35} & \multicolumn{1}{c}{57.55}&\multicolumn{1}{c}{\textbf{58.54}}\\\hline\hline
\multicolumn{1}{l}{\textbf{QNLI}} & \multicolumn{5}{c}{\textbf{}} & \multicolumn{3}{c}{\textbf{ }} \\ \hline
\multicolumn{1}{l|}{Names, Career/Family} &\multicolumn{1}{c|}{0.26} &\multicolumn{1}{l}{0.05}&\multicolumn{1}{l}{0.10}&\multicolumn{1}{c}{\textbf{0.01}}&\multicolumn{1}{c|}{0.02}&  \multicolumn{1}{c|}{0.04} & \multicolumn{1}{c}{0.38}&\multicolumn{1}{c}{\textbf{0.17}}\\
\multicolumn{1}{l|}{Terms, Career/Family} & \multicolumn{1}{c|}{0.15} &\multicolumn{1}{l}{\textbf{0.004}}&\multicolumn{1}{l}{0.20}& \multicolumn{1}{c}{0.13}&\multicolumn{1}{c|}{0.04}&  \multicolumn{1}{c|}{0.22} & \multicolumn{1}{c}{0.10}&\multicolumn{1}{c}{\textbf{0.04}}\\
\multicolumn{1}{l|}{Terms, Math/Art} &\multicolumn{1}{c|}{0.58} &\multicolumn{1}{l}{\textbf{0.08}}&\multicolumn{1}{l}{0.32}& \multicolumn{1}{c}{0.30}&\multicolumn{1}{c|}{\textbf{0.08}}&  \multicolumn{1}{c|}{0.53} & \multicolumn{1}{c}{0.16}&\multicolumn{1}{c}{\textbf{0.09}}\\
\multicolumn{1}{l|}{Names, Math/Art} &\multicolumn{1}{c|}{0.58} &\multicolumn{1}{l}{0.62}&\multicolumn{1}{l}{0.28}& \multicolumn{1}{c}{0.23}&\multicolumn{1}{c|}{\textbf{0.16}}&  \multicolumn{1}{c|}{0.48} & \multicolumn{1}{c}{0.06}&\multicolumn{1}{c}{\textbf{0.03}}\\
\multicolumn{1}{l|}{Terms, Science/Art} &\multicolumn{1}{c|}{0.08}&\multicolumn{1}{l}{0.71}&\multicolumn{1}{l}{0.24} & \multicolumn{1}{c}{0.25}&\multicolumn{1}{c|}{\textbf{0.21}}&  \multicolumn{1}{c|}{0.47} & \multicolumn{1}{c}{0.57}&\multicolumn{1}{c}{\textbf{0.53}}\\
\multicolumn{1}{l|}{Names, Science/Art} &\multicolumn{1}{c|}{0.52} &\multicolumn{1}{l}{0.44}&\multicolumn{1}{l}{0.16}& \multicolumn{1}{c}{0.15}&\multicolumn{1}{c|}{\textbf{0.04}}&  \multicolumn{1}{c|}{0.36} & \multicolumn{1}{c}{0.47}&\multicolumn{1}{c}{0.52}\\\hline
\multicolumn{1}{l|}{Avg. Abs. e-size} &\multicolumn{1}{c|}{0.365} &\multicolumn{1}{l}{0.321}&\multicolumn{1}{l}{0.222}& \multicolumn{1}{c}{0.178}&\multicolumn{1}{c|}{\textbf{0.091}}&  \multicolumn{1}{c|}{0.350} & \multicolumn{1}{c}{0.290}&\multicolumn{1}{c}{\textbf{0.230}}\\\hline
\multicolumn{1}{l|}{Accuracy} &\multicolumn{1}{c|}{91.30} &\multicolumn{1}{l}{90.60}&\multicolumn{1}{l}{90.80}& \multicolumn{1}{c}{\textbf{91.61}}&\multicolumn{1}{c|}{91.32}&  \multicolumn{1}{c|}{92.03} & \multicolumn{1}{c}{\textbf{92.51}}&\multicolumn{1}{c}{92.09}\\\hline

\end{tabular}}
\caption{ Comparing the debiasing performance of Gender-tuning and two state-of-the-art baselines.  First six rows measure binary SEAT effect size (e-size; lower is better) for sentence-level tests from \cite{caliskan2017semantics}. The seventh row presents the average absolute e-size. The eighth row shows the classification accuracy on downstream tasks. The Gender-tuning-random masks the input example randomly (not only gender-words). Gender-tuning gains the lowest average bias in both models and all datasets.}
\label{tab:table2}
\end{table*}

\begin{table*}[t]
\centering
\resizebox{13.8cm}{!}{
\begin{tabular}{ l| l| l | l}
\hline

 \multicolumn{1}{l}{Training input } & \multicolumn{1}{l}{Perturbed } & \multicolumn{1}{l}{Type} & \multicolumn{1}{c}{Label} \\ \hline
 \multicolumn{1}{l|}{with [\textbf{his}] usual intelligence and subtlety. } & \multicolumn{1}{l|}{with [\textbf{\color{blue}{the}]} usual intelligence and subtlety. } & \multicolumn{1}{l|}{neutral} & \multicolumn{1}{c}{1} \\ 
 \multicolumn{1}{l|}{ } & \multicolumn{1}{l|}{} & \multicolumn{1}{l|}{} & \multicolumn{1}{l}{} \\
 \multicolumn{1}{l|}{by casting an [\textbf{actress}] whose face projects } & \multicolumn{1}{l|}{by casting an [\textbf{\color{blue}{image}}] whose face projects} & \multicolumn{1}{l|}{} & \multicolumn{1}{l}{} \\
 \multicolumn{1}{l|}{  that [\textbf{woman}] 's doubts and yearnings ,} & \multicolumn{1}{l|}{ that [\textbf{\color{blue}{person}}] 's doubts and yearnings ,} & \multicolumn{1}{l|}{neutral} & \multicolumn{1}{c}{1} \\ 
 \multicolumn{1}{l|}{ it succeeds.} & \multicolumn{1}{l|}{ it succeeds.} & \multicolumn{1}{l|}{} & \multicolumn{1}{l}{} \\ 
 \multicolumn{1}{l|}{ } & \multicolumn{1}{l|}{} & \multicolumn{1}{l|}{} & \multicolumn{1}{l}{} \\
\multicolumn{1}{l|}{certainly has a new career ahead of [\textbf{him}] if } & \multicolumn{1}{l|}{certainly has a new career ahead of [\textbf{\color{blue}{her}}] if } & \multicolumn{1}{l|}{convert-gender} & \multicolumn{1}{c}{1} \\ 
 \multicolumn{1}{l|}{[\textbf{he}] so chooses. } & \multicolumn{1}{l|}{[\textbf{\color{blue}{she}}] so chooses.} & \multicolumn{1}{l|}{} & \multicolumn{1}{l}{} \\
 \multicolumn{1}{l|}{ } & \multicolumn{1}{l|}{} & \multicolumn{1}{l|}{} & \multicolumn{1}{l}{} \\
 \multicolumn{1}{l|}{by [\textbf{men}] of marginal intelligence , with } & \multicolumn{1}{l|}{by [\textbf{\color{blue}{people}}] of marginal intelligence , with } & \multicolumn{1}{l|}{neutral} & \multicolumn{1}{c}{0} \\
 \multicolumn{1}{l|}{ reactionary ideas. } & \multicolumn{1}{l|}{reactionary ideas.} & \multicolumn{1}{l|}{} & \multicolumn{1}{l}{} \\
\multicolumn{1}{l|}{ } & \multicolumn{1}{l|}{} & \multicolumn{1}{l|}{} & \multicolumn{1}{l}{} \\
\multicolumn{1}{l|}{why this distinguished [\textbf{actor}] would stoop so low.} & \multicolumn{1}{l|}{why this distinguished [\textbf{\color{blue}{man}}] would stoop so low.} & \multicolumn{1}{l|}{same-gender} & \multicolumn{1}{c}{0} \\
 \multicolumn{1}{l|}{ } & \multicolumn{1}{l|}{} & \multicolumn{1}{l|}{} & \multicolumn{1}{l}{} \\
 \multicolumn{1}{l|}{it is very awful - - and oozing with creepy [\textbf{men}].} & \multicolumn{1}{l|}{it is very awful - - and oozing with creepy [{\color{blue}{UNK}}] .} & \multicolumn{1}{l|}{deleting} & \multicolumn{1}{c}{0} \\
 \multicolumn{1}{l|}{ } & \multicolumn{1}{l|}{} & \multicolumn{1}{l|}{} & \multicolumn{1}{l}{} \\
 \multicolumn{1}{l|}{Proves once again [\textbf{he}] hasn't lost.} & \multicolumn{1}{l|}{Proves once again [\textbf{\color{blue}{he}}]  hasn't lost .} & \multicolumn{1}{l|}{identical} & \multicolumn{1}{c}{1} \\\hline
\end{tabular}}
\caption{The illustration of the different types of perturbation outputs generated by Gender-tuning and their ground-truth label. }
\label{tab:table1}
\end{table*}

\begin{table*}[t]
\centering

\resizebox{13.9cm}{!}{
\begin{tabular}{l|c cc c||c c c c}
\hline

\multicolumn{1}{l}{\textbf{SST-2}} & \multicolumn{4}{c}{\textbf{BERT }} & \multicolumn{4}{c}{\textbf{RoBERTa }} \\ \hline
\multicolumn{1}{l|}{}&  \multicolumn{1}{l|}{Origin} & \multicolumn{1}{l}{Gender-tuning/} &\multicolumn{1}{l}{Gender-tuning/}&\multicolumn{1}{l|}{Gender-tuning}&\multicolumn{1}{l|}{Origin} &\multicolumn{1}{l}{Gender-tuning/} & \multicolumn{1}{l}{Gender-tuning/}&\multicolumn{1}{l}{Gender-tuning}\\ 
\multicolumn{1}{l|}{}&  \multicolumn{1}{l|}{} & \multicolumn{1}{l}{no-joint-train} &\multicolumn{1}{l}{no-joint-loss}&\multicolumn{1}{c|}{(ours)}&\multicolumn{1}{l|}{} &\multicolumn{1}{l}{no-joint-train} & \multicolumn{1}{l}{no-joint-loss}&\multicolumn{1}{c}{(ours)}\\ \hline
\multicolumn{1}{l|}{Names, Career/Family} &\multicolumn{1}{c|}{0.03} &\multicolumn{1}{c}{0.22} & \multicolumn{1}{c}{0.16}&\multicolumn{1}{c|}{\textbf{0.03}}&  \multicolumn{1}{c|}{0.07} &\multicolumn{1}{c}{0.18} & \multicolumn{1}{c}{0.62}&\multicolumn{1}{c}{\textbf{0.14}}\\
\multicolumn{1}{l|}{Terms, Career/Family} &\multicolumn{1}{c|}{0.01} & \multicolumn{1}{c}{0.31} &\multicolumn{1}{c}{0.37}&\multicolumn{1}{c|}{\textbf{0.16}}&  \multicolumn{1}{c|}{0.33} &\multicolumn{1}{c}{0.09} & \multicolumn{1}{c}{0.41}&\multicolumn{1}{c}{\textbf{0.01}}\\
\multicolumn{1}{l|}{Terms, Math/Art} &\multicolumn{1}{c|}{0.21} & \multicolumn{1}{c}{0.75} &\multicolumn{1}{c}{0.49}&\multicolumn{1}{c|}{\textbf{0.39}}&  \multicolumn{1}{c|}{1.32} &\multicolumn{1}{c}{0.99} & \multicolumn{1}{c}{1.02}&\multicolumn{1}{c}{\textbf{0.57}}\\
\multicolumn{1}{l|}{Names, Math/Art} &\multicolumn{1}{c|}{1.15} & \multicolumn{1}{c}{0.55} &\multicolumn{1}{c}{0.56}&\multicolumn{1}{c|}{\textbf{0.31}}&  \multicolumn{1}{c|}{1.34} &\multicolumn{1}{c}{\textbf{0.92}} & \multicolumn{1}{c}{0.97}&\multicolumn{1}{c}{1.11}\\
\multicolumn{1}{l|}{Terms, Science/Art} &\multicolumn{1}{c|}{0.10} & \multicolumn{1}{c}{0.01} &\multicolumn{1}{c}{0.32}&\multicolumn{1}{c|}{\textbf{0.07}}&  \multicolumn{1}{c|}{0.25} &\multicolumn{1}{c}{0.76} & \multicolumn{1}{c}{\textbf{0.00}}&\multicolumn{1}{c}{0.47}\\
\multicolumn{1}{l|}{Names, Science/Art} &\multicolumn{1}{c|}{0.22} & \multicolumn{1}{c}{\textbf{0.07}} &\multicolumn{1}{c}{0.47}&\multicolumn{1}{c|}{0.10}&  \multicolumn{1}{c|}{0.47} &\multicolumn{1}{c}{0.76} & \multicolumn{1}{c}{0.56}&\multicolumn{1}{c}{\textbf{0.47}}\\\hline
\multicolumn{1}{l|}{Avg. Abs. e-size} &\multicolumn{1}{c|}{0.291} & \multicolumn{1}{c}{0.318} &\multicolumn{1}{c}{0.395}&\multicolumn{1}{c|}{\textbf{0.176}}&  \multicolumn{1}{c|}{0.630} &\multicolumn{1}{c}{0.616} & \multicolumn{1}{c}{0.596}&\multicolumn{1}{c}{\textbf{0.461}}\\\hline
\multicolumn{1}{l|}{Accuracy} &\multicolumn{1}{c|}{91.97} & \multicolumn{1}{c}{\textbf{92.88}} &\multicolumn{1}{c}{92.66}&\multicolumn{1}{c|}{92.10}&  \multicolumn{1}{c|}{93.57 } &\multicolumn{1}{c}{\textbf{94.38}} & \multicolumn{1}{c}{92.54}&\multicolumn{1}{c}{93.69}\\\hline\hline
\multicolumn{1}{l}{\textbf{CoLA}} & \multicolumn{4}{c}{\textbf{ }} & \multicolumn{4}{c}{\textbf{ }} \\ \hline
\multicolumn{1}{l|}{Names, Career/Family} &\multicolumn{1}{c|}{0.09} & \multicolumn{1}{c}{0.37} &\multicolumn{1}{c}{\textbf{0.04}}&\multicolumn{1}{c|}{0.09}&  \multicolumn{1}{c|}{0.29} &\multicolumn{1}{c}{0.07} & \multicolumn{1}{c}{0.16}&\multicolumn{1}{c}{\textbf{0.05}}\\
\multicolumn{1}{l|}{Terms, Career/Family} &\multicolumn{1}{c|}{0.19} & \multicolumn{1}{c}{0.06} &\multicolumn{1}{c}{0.11}&\multicolumn{1}{c|}{\textbf{0.03}}&  \multicolumn{1}{c|}{0.26} &\multicolumn{1}{c}{0.16} & \multicolumn{1}{c}{0.11}&\multicolumn{1}{c}{\textbf{0.00}}\\
\multicolumn{1}{l|}{Terms, Math/Art} &\multicolumn{1}{c|}{0.26} & \multicolumn{1}{c}{0.89} &\multicolumn{1}{c}{0.96}&\multicolumn{1}{c|}{\textbf{0.08}}&  \multicolumn{1}{c|}{0.06} &\multicolumn{1}{c}{0.41} & \multicolumn{1}{c}{0.29}&\multicolumn{1}{c}{\textbf{0.15}}\\
\multicolumn{1}{l|}{Names, Math/Art} &\multicolumn{1}{c|}{0.15} &\multicolumn{1}{c}{1.03} & \multicolumn{1}{c}{0.82}&\multicolumn{1}{c|}{\textbf{0.24}}&  \multicolumn{1}{c|}{0.06} &\multicolumn{1}{c}{0.22} & \multicolumn{1}{c}{0.87}&\multicolumn{1}{c}{\textbf{0.07}}\\
\multicolumn{1}{l|}{Terms, Science/Art} &\multicolumn{1}{c|}{0.42} & \multicolumn{1}{c}{0.47} &\multicolumn{1}{c}{0.19}&\multicolumn{1}{c|}{\textbf{0.07}}&  \multicolumn{1}{c|}{0.32} &\multicolumn{1}{c}{\textbf{0.42}} & \multicolumn{1}{c}{0.80}&\multicolumn{1}{c}{0.70}\\
\multicolumn{1}{l|}{Names, Science/Art} &\multicolumn{1}{c|}{0.03} &\multicolumn{1}{c}{0.49} & \multicolumn{1}{c}{0.32}&\multicolumn{1}{c|}{\textbf{0.07}}&  \multicolumn{1}{c|}{0.27} &\multicolumn{1}{c}{0.36} & \multicolumn{1}{c}{0.88}&\multicolumn{1}{c}{\textbf{0.03}}\\\hline
\multicolumn{1}{l|}{Avg. Abs. e-size} &\multicolumn{1}{c|}{0.181} & \multicolumn{1}{c}{0.551} &\multicolumn{1}{c}{0.406}&\multicolumn{1}{c|}{\textbf{0.096}}&  \multicolumn{1}{c|}{0.210} &\multicolumn{1}{c}{0.273} & \multicolumn{1}{c}{0.518}&\multicolumn{1}{c}{\textbf{0.166}}\\\hline
\multicolumn{1}{l|}{Accuracy} &\multicolumn{1}{c|}{56.51} & \multicolumn{1}{c}{56.32} &\multicolumn{1}{c}{\textbf{56.70}}&\multicolumn{1}{c|}{56.60}& \multicolumn{1}{c|}{ 57.35} & \multicolumn{1}{c}{\textbf{62.11}} & \multicolumn{1}{c}{57.27}&\multicolumn{1}{c}{58.54}\\\hline\hline
\multicolumn{1}{l}{\textbf{QNLI}} & \multicolumn{4}{c}{\textbf{ }} & \multicolumn{4}{c}{\textbf{ }} \\ \hline
\multicolumn{1}{l|}{Names, Career/Family} &\multicolumn{1}{c|}{0.26} &\multicolumn{1}{c}{0.03} & \multicolumn{1}{c}{0.15}&\multicolumn{1}{c|}{\textbf{0.02}}&  \multicolumn{1}{c|}{0.04} &\multicolumn{1}{c}{\textbf{0.12}} & \multicolumn{1}{c}{0.14}&\multicolumn{1}{c}{0.17}\\
\multicolumn{1}{l|}{Terms, Career/Family} &\multicolumn{1}{c|}{0.15} & \multicolumn{1}{c}{0.20} &\multicolumn{1}{c}{0.41}&\multicolumn{1}{c|}{\textbf{0.04}}&  \multicolumn{1}{c|}{0.22} &\multicolumn{1}{c}{0.31} & \multicolumn{1}{c}{0.11}&\multicolumn{1}{c}{\textbf{0.04}}\\
\multicolumn{1}{l|}{Terms, Math/Art} &\multicolumn{1}{c|}{0.58} & \multicolumn{1}{c}{0.47} &\multicolumn{1}{c}{\textbf{0.03}}&\multicolumn{1}{c|}{0.08}&  \multicolumn{1}{c|}{0.53} & \multicolumn{1}{c}{0.50} & \multicolumn{1}{c}{0.62}&\multicolumn{1}{c}{\textbf{0.09}}\\
\multicolumn{1}{l|}{Names, Math/Art} &\multicolumn{1}{c|}{0.58} & \multicolumn{1}{c}{0.94} &\multicolumn{1}{c}{\textbf{0.04}}&\multicolumn{1}{c|}{0.16}&  \multicolumn{1}{c|}{0.48} &\multicolumn{1}{c}{0.38} & \multicolumn{1}{c}{0.42}&\multicolumn{1}{c}{\textbf{0.03}}\\
\multicolumn{1}{l|}{Terms, Science/Art} &\multicolumn{1}{c|}{0.08} & \multicolumn{1}{c}{\textbf{0.12}} &\multicolumn{1}{c}{0.27}&\multicolumn{1}{c|}{0.21}&  \multicolumn{1}{c|}{0.47} & \multicolumn{1}{c}{\textbf{0.25}} & \multicolumn{1}{c}{0.50}&\multicolumn{1}{c}{0.53}\\
\multicolumn{1}{l|}{Names, Science/Art} &\multicolumn{1}{c|}{0.52} & \multicolumn{1}{c}{0.54} &\multicolumn{1}{c}{0.11}&\multicolumn{1}{c|}{\textbf{0.04}}&  \multicolumn{1}{c|}{0.36} & \multicolumn{1}{c}{\textbf{0.03}} & \multicolumn{1}{c}{0.20}&\multicolumn{1}{c}{0.52}\\\hline
\multicolumn{1}{l|}{Avg. Abs. e-size} &\multicolumn{1}{c|}{0.365} & \multicolumn{1}{c}{0.383} &\multicolumn{1}{c}{0.168}&\multicolumn{1}{c|}{\textbf{0.091}}&  \multicolumn{1}{c|}{0.350} & \multicolumn{1}{c}{0.265} & \multicolumn{1}{c}{0.331}&\multicolumn{1}{c}{\textbf{0.230}}\\\hline
\multicolumn{1}{l|}{Accuracy} &\multicolumn{1}{c|}{91.30} & \multicolumn{1}{c}{\textbf{91.57}} &\multicolumn{1}{c}{91.28}&\multicolumn{1}{c|}{91.32}&  \multicolumn{1}{c|}{92.03 } &\multicolumn{1}{c}{\textbf{92.58}} & \multicolumn{1}{c}{91.69}&\multicolumn{1}{c}{92.09}\\\hline

\end{tabular}}
\caption{ Comparing the debiasing performance of two ablation experiments and Gender-tuning (ours) on three downstream task datasets. The results show that Gender-tuning achieved the least average bias score and consistently improved the classification accuracy.}
\label{tab:table3}
\end{table*} 


Suppose the MLM predicts an incorrect token. For instance, the example: "the film affirms the power of the [actress]" changes to   "the film affirms the power of the [trauma]". In this example, the predicted word [trauma] is a non-related gender-word that raises perturbation-loss value ($\mathcal{L}_{perturb}$ > $0$). In this case, even if fine-tuning classifies the perturbed example correctly, joint-loss is still big enough to force Gender-tuning to continue training. 

Also, suppose Gender-tuning creates social gender bias through gender perturbation. For instance, the example: "angry black [actor]" changes to  "angry black [woman]" that "woman" and "actor" are not close semantically that raises perturbation-loss value ($\mathcal{L}_{perturb}$ > $0$). In this case, the output of the fine-tuning might be correct ($\mathcal{L}_{fine-tuning}$ $\approx{0})$  due to the PLMs' learned biases ("angry black woman" is a known gender/race bias). However, due to the big value of perturbation-loss, the join-loss is big enough to override fine-tuning results and forces Gender-tuning to continue training.

Moreover, we observed that sometimes example perturbation changes the concept/label of training examples. For instance,  the input: "[He] is an excellent [actor] (label: positive)" changes to "[She] is a wonderful [murderer] (label: positive)", and fine-tuning classification output is correct ($\mathcal{L}_{fine-tuning} \approx{0})$. In this example, the predicted word [murderer] is conceptually far from gender-related words [actor]. So, perturbation loss becomes significant, which creates a big value for joint-loss to force Gender-tuning to continue training.  
Finally, we found examples that MLM replaces the gender-word with the [UNK] token. In these examples, the perturbation-loss is close to zero ($\mathcal{L}_{perturb} \approx{0})$ and the output of the fine-tuning classifier is incorrect ($\mathcal{L}_{fine-tuning}$ > $0$). In this case, the joint-loss is big enough to continue training and provide a new chance for MLM to predict a meaningful token instead of a [UNK]. More analysis of our perturbation strategy can be found in Section \ref{sec:4.1}
 and Table \ref{tab:table1}.
 
\section{Experimental Setup}
\label{section:3}
To evaluate our proposed method, we conduct experiments by following the evaluation process of the two state-of-the-art baselines (Sent-D and FairF) such as the bias evaluation metric (SEAT), applied PLMs, and downstream tasks' datasets. (Details of the baselines, bias evaluation metric, PLMs, datasets, and hyperparameters are presented in Appendix \ref{sec:A})

We report the SEAT effect size (e-size), average absolute e-size, and classification accuracy on downstream tasks for three different setups: 1) \textbf{Origin}: fine-tuning the PLMs on the downstream task datasets using huggingface transformers code \cite{wolf-etal-2020-transformers}. 2) \textbf{Gender-tuning}-random: instead of replacing the gender-words in an training example, Gender-tuning-random replaces a certain percentage of an input tokens randomly (5\% of each input sequence). 3) \textbf{Gender-tuning}: the proposed method. We used the same hyperparameter for all three setups for a fair comparison. 

\section{Results and Discussion}
Table \ref{tab:table2} illustrates  SEAT absolute effect size (e-size) (lower is better) on sentence templates of Terms/Names under different gender domains provided by \citep{caliskan2017semantics},  average absolute e-size (lower is better), and classification accuracy on downstream tasks (higher is better) for three experiment setups (Section \ref{section:3}) and two state-of-the-art baselines. The results show that Gender-tuning outperforms the baselines regarding the average absolute effect size for both PLMs on all datasets. Also, in contrast with the baselines, Gender-tuning improves the accuracy of both PLMs on all downstream tasks. It shows that the proposed method preserves the useful semantic information of the training data after debiasing. 
The Gender-tuning-random results show an inconsistent effect on the bias scores. Although Gender-tuning-random improves the PLMs' accuracy on the downstream tasks,  it significantly magnifies the bias score in the BERT model on SST-2 and CoLA. Also,  it slightly reduces the average bias score in the RoBERTa on all datasets and in BERT on the QNLI.

\subsection{Perturbation Analysis}
\label{sec:4.1}

The PLMs achieved state-of-the-art performance on the downstream tasks datasets by applying the MLM for the example perturbation in pre-training phase.
Thus we hypothesize that the MLM can generate realistic gender-perturbed examples that can considerably modify the gender relation between the input tokens without affecting the label. 
However, there is a concern that the pre-trained MLM transfers the gender bias through the perturbation process. 

To address this concern, we investigate the predicted tokens that the pre-trained MLM replaces with the gender-words. We randomly select 300 examples from training dataset including 150 examples with feminine words and 150 examples with masculine words. Based on these 300 examples, we observe five types of perturbation as shown through some examples in Table \ref{tab:table1}:
\begin{itemize}
\item  \textbf{Neutral}; replace the gender-words with neutral word such as people, they, their, and etc.
\item  \textbf{Convert-gender}; replace the gender-words with opposite gender. the word "he" change to "she". 
\item  \textbf{Same-gender}; replace the gender-words with the same gender. change the word "man" to "boy". 
\item  \textbf{Deleting}; replace the gender-words with unknown token ([UNK]). In 300 examples, it only happens when there are several masked tokens.
\item  \textbf{Identical}; replace the gender-word with itself. It mostly happens when there is only one gender-word.
\end{itemize} 

In our investigation with 300 examples, we had 46\% Neutral, 29\% Identical, 17\% Convert-gender, 7\% Same-gender, and  1\% Deleting perturbation. As illustrated in Table \ref{tab:table1}, Gender-tuning does not make a meaningful change in identical and same-gender perturbation. These examples likely conform to the gender biases in the MLM. Suppose identical, or same-gender perturbation gets the correct output from the perturbation process  ($\mathcal{L}_{perturb.}$ $\approx{0}$). 
In this case, the only way to learn the biases in the MLM is to get the correct output from fine-tuning step and joint-loss close to zero. This issue stops the MLM and fine-tuning model from further update. However, joint-loss plays an essential role in alleviating learning gender bias from identical and same-gender perturbations.  

To clarify the role of joint-loss in overcoming above problem, we investigated fine-tuning output on identical and same-gender perturbations. We observed that fine-tuning gets the incorrect output from 60\% of the identical and 75\% of the same-gender perturbation. Thus these examples return to training iteration because their joint-loss is large enough to update the language models and perform a new training iteration. New training iteration means re-perturbing and re-fine-tuning result on these examples. Therefore,  training based on both training steps' loss and computing  joint-loss persistently prevents learning from gender bias in MLM as well as the PLM.

\section{Ablation}
We conduct the ablation experiments to demonstrate the effectiveness of Gender-tuning components, including 1) joint-training process and 2) joint-loss in Gender-tuning's debiasing performance (Table \ref{tab:table3}). The experiments are as follows:
1)  \textbf{Gender-tuning$_{no-joint-training}$}: first we used MLM to train the PLM through the gender-word perturbation on downstream task datasets. Then we fine-tuned the PLM on the downstream task dataset. 2) \textbf{Gender-tuning$_{no-joint-loss}$}: we train Gender-tuning  based on only fine-tuning loss. 

In both PLMs, results illustrate that Gender-tuning is more effective for reducing the average gender bias than in two ablation experiments. The two ablation experiments magnify the bias scores noticeably, while Gender-tuning gains the smallest SEAT absolute effect size, especially in the BERT model. 
Results also show that the ablation experiment setups that do not benefit from joint-loss cannot update the MLM and PLM when the output of the fine-tuning classification is correct ($\mathcal{L}_{fine-tuning}$ $\approx{0}$), even though the correct output likely bases on the gender biases in the PLMs.  


\section{Conclusion}
We propose a novel approach for debiasing PLMs through fine-tuning on downstream tasks' datasets. The proposed method is an aggregation of bias-word perturbation using MLM and fine-tuning classification. In this study, we evaluated our proposed method on gender biases and named it \emph{Gender-tuning}. 
Comprehensive experiments prove that Gender-tuning outperforms two state-of-the-art debiasing methods while improving the performance of the PLMs on downstream tasks. 
The key advantage of our approach is using the fine-tuning setting that allows the training process to be carried out without needing additional training processes or datasets. Also,  it makes Gender-tuning a plug-and-play debiasing tool deployable to any PLMs.

\section{Limitation}
Although Gender-tuning succeeds in reducing the gender bias scores in the pre-trained language models, there are some limitations to performing debiasing. Gender-tuning only works on gender-related words list. Thus Gender-tuning cannot cover the probable gender biases that do not exist in its' list. We defer the gender-related word list modification to future research. All our experiments ran on English language texts with English gender-word morphology.
\bibliography{anthology,custom}

\begin{thebibliography}{33}
\expandafter\ifx\csname natexlab\endcsname\relax\def\natexlab#1{#1}\fi

\bibitem[{Barikeri et~al.(2021)Barikeri, Lauscher, Vuli{\'c}, and
  Glava{\v{s}}}]{barikeri2021redditbias}
Soumya Barikeri, Anne Lauscher, Ivan Vuli{\'c}, and Goran Glava{\v{s}}. 2021.
\newblock Redditbias: A real-world resource for bias evaluation and debiasing
  of conversational language models.
\newblock \emph{arXiv preprint arXiv:2106.03521}.

\bibitem[{Bender et~al.(2021)Bender, Gebru, McMillan-Major, and
  Shmitchell}]{bender2021dangers}
Emily~M Bender, Timnit Gebru, Angelina McMillan-Major, and Shmargaret
  Shmitchell. 2021.
\newblock On the dangers of stochastic parrots: Can language models be too big?
\newblock In \emph{Proceedings of the 2021 ACM Conference on Fairness,
  Accountability, and Transparency}, pages 610--623.

\bibitem[{Bolukbasi et~al.(2016)Bolukbasi, Chang, Zou, Saligrama, and
  Kalai}]{bolukbasi2016man}
Tolga Bolukbasi, Kai-Wei Chang, James~Y Zou, Venkatesh Saligrama, and Adam~T
  Kalai. 2016.
\newblock Man is to computer programmer as woman is to homemaker? debiasing
  word embeddings.
\newblock \emph{Advances in neural information processing systems},
  29:4349--4357.

\bibitem[{Bordia and Bowman(2019)}]{bordia2019identifying}
Shikha Bordia and Samuel~R Bowman. 2019.
\newblock Identifying and reducing gender bias in word-level language models.
\newblock \emph{NAACL HLT 2019}, page~7.

\bibitem[{Brunet et~al.(2019)Brunet, Alkalay-Houlihan, Anderson, and
  Zemel}]{brunet2019understanding}
Marc-Etienne Brunet, Colleen Alkalay-Houlihan, Ashton Anderson, and Richard
  Zemel. 2019.
\newblock Understanding the origins of bias in word embeddings.
\newblock In \emph{International conference on machine learning}, pages
  803--811. PMLR.

\bibitem[{Caliskan et~al.(2017)Caliskan, Bryson, and
  Narayanan}]{caliskan2017semantics}
Aylin Caliskan, Joanna~J Bryson, and Arvind Narayanan. 2017.
\newblock Semantics derived automatically from language corpora contain
  human-like biases.
\newblock \emph{Science}, 356(6334):183--186.

\bibitem[{Chen et~al.(2020)Chen, Kornblith, Norouzi, and
  Hinton}]{chen2020simple}
Ting Chen, Simon Kornblith, Mohammad Norouzi, and Geoffrey Hinton. 2020.
\newblock A simple framework for contrastive learning of visual
  representations.
\newblock In \emph{International conference on machine learning}, pages
  1597--1607. PMLR.

\bibitem[{Cheng et~al.(2021)Cheng, Hao, Yuan, Si, and Carin}]{cheng2021fairfil}
Pengyu Cheng, Weituo Hao, Siyang Yuan, Shijing Si, and Lawrence Carin. 2021.
\newblock Fairfil: Contrastive neural debiasing method for pretrained text
  encoders.
\newblock In \emph{International Conference on Learning Representations}.

\bibitem[{Clark et~al.(2020)Clark, Luong, Le, and Manning}]{clark2020electra}
Kevin Clark, Minh-Thang Luong, Quoc~V Le, and Christopher~D Manning. 2020.
\newblock Electra: Pre-training text encoders as discriminators rather than
  generators.

\bibitem[{Dev et~al.(2020)Dev, Li, Phillips, and Srikumar}]{dev2020measuring}
Sunipa Dev, Tao Li, Jeff~M Phillips, and Vivek Srikumar. 2020.
\newblock On measuring and mitigating biased inferences of word embeddings.
\newblock In \emph{Proceedings of the AAAI Conference on Artificial
  Intelligence}, volume~34, pages 7659--7666.

\bibitem[{Devlin et~al.(2019)Devlin, Chang, Lee, and
  Toutanova}]{devlin2019bert}
Jacob Devlin, Ming-Wei Chang, Kenton Lee, and Kristina Toutanova. 2019.
\newblock Bert: Pre-training of deep bidirectional transformers for language
  understanding.
\newblock In \emph{Proceedings of the 2019 Conference of the North American
  Chapter of the Association for Computational Linguistics: Human Language
  Technologies, Volume 1 (Long and Short Papers)}, pages 4171--4186.

\bibitem[{Dinan et~al.(2020)Dinan, Fan, Williams, Urbanek, Kiela, and
  Weston}]{dinan2020queens}
Emily Dinan, Angela Fan, Adina Williams, Jack Urbanek, Douwe Kiela, and Jason
  Weston. 2020.
\newblock Queens are powerful too: Mitigating gender bias in dialogue
  generation.
\newblock In \emph{Proceedings of the 2020 Conference on Empirical Methods in
  Natural Language Processing (EMNLP)}, pages 8173--8188.

\bibitem[{Dixon et~al.(2018)Dixon, Li, Sorensen, Thain, and
  Vasserman}]{dixon2018measuring}
Lucas Dixon, John Li, Jeffrey Sorensen, Nithum Thain, and Lucy Vasserman. 2018.
\newblock Measuring and mitigating unintended bias in text classification.
\newblock In \emph{Proceedings of the 2018 AAAI/ACM Conference on AI, Ethics,
  and Society}, pages 67--73.

\bibitem[{Gehman et~al.(2020)Gehman, Gururangan, Sap, Choi, and
  Smith}]{gehman2020realtoxicityprompts}
Samuel Gehman, Suchin Gururangan, Maarten Sap, Yejin Choi, and Noah~A Smith.
  2020.
\newblock Realtoxicityprompts: Evaluating neural toxic degeneration in language
  models.
\newblock In \emph{Proceedings of the 2020 Conference on Empirical Methods in
  Natural Language Processing: Findings}, pages 3356--3369.

\bibitem[{Kaneko and Bollegala(2019)}]{kaneko2019gender}
Masahiro Kaneko and Danushka Bollegala. 2019.
\newblock Gender-preserving debiasing for pre-trained word embeddings.
\newblock In \emph{Proceedings of the 57th Annual Meeting of the Association
  for Computational Linguistics}, pages 1641--1650.

\bibitem[{Kurita et~al.(2019)Kurita, Vyas, Pareek, Black, and
  Tsvetkov}]{kurita2019measuring}
Keita Kurita, Nidhi Vyas, Ayush Pareek, Alan~W Black, and Yulia Tsvetkov. 2019.
\newblock Measuring bias in contextualized word representations.
\newblock In \emph{Proceedings of the First Workshop on Gender Bias in Natural
  Language Processing}, pages 166--172.

\bibitem[{Liang et~al.(2020)Liang, Li, Zheng, Lim, Salakhutdinov, and
  Morency}]{liang2020towards}
Paul~Pu Liang, Irene~Mengze Li, Emily Zheng, Yao~Chong Lim, Ruslan
  Salakhutdinov, and Louis-Philippe Morency. 2020.
\newblock Towards debiasing sentence representations.
\newblock In \emph{Proceedings of the 58th Annual Meeting of the Association
  for Computational Linguistics}.

\bibitem[{Liu et~al.(2019)Liu, Ott, Goyal, Du, Joshi, Chen, Levy, Lewis,
  Zettlemoyer, and Stoyanov}]{liu2019roberta}
Yinhan Liu, Myle Ott, Naman Goyal, Jingfei Du, Mandar Joshi, Danqi Chen, Omer
  Levy, Mike Lewis, Luke Zettlemoyer, and Veselin Stoyanov. 2019.
\newblock Roberta: A robustly optimized bert pretraining approach.
\newblock \emph{arXiv preprint arXiv:1907.11692}.

\bibitem[{Manzini et~al.(2019)Manzini, Chong, Black, and
  Tsvetkov}]{manzini2019black}
Thomas Manzini, Lim~Yao Chong, Alan~W Black, and Yulia Tsvetkov. 2019.
\newblock Black is to criminal as caucasian is to police: Detecting and
  removing multiclass bias in word embeddings.
\newblock In \emph{Proceedings of the 2019 Conference of the North American
  Chapter of the Association for Computational Linguistics: Human Language
  Technologies, Volume 1 (Long and Short Papers)}, pages 615--621.

\bibitem[{May et~al.(2019)May, Wang, Bordia, Bowman, and
  Rudinger}]{may2019measuring}
Chandler May, Alex Wang, Shikha Bordia, Samuel~R Bowman, and Rachel Rudinger.
  2019.
\newblock On measuring social biases in sentence encoders.
\newblock In \emph{NAACL-HLT (1)}.

\bibitem[{Meade et~al.(2022)Meade, Poole-Dayan, and Reddy}]{meade2022empirical}
Nicholas Meade, Elinor Poole-Dayan, and Siva Reddy. 2022.
\newblock An empirical survey of the effectiveness of debiasing techniques for
  pre-trained language models.
\newblock In \emph{Proceedings of the 60th Annual Meeting of the Association
  for Computational Linguistics (Volume 1: Long Papers)}, pages 1878--1898.

\bibitem[{Raffel et~al.(2020)Raffel, Shazeer, Roberts, Lee, Narang, Matena,
  Zhou, Li, and Liu}]{raffel2020exploring}
Colin Raffel, Noam Shazeer, Adam Roberts, Katherine Lee, Sharan Narang, Michael
  Matena, Yanqi Zhou, Wei Li, and Peter~J Liu. 2020.
\newblock Exploring the limits of transfer learning with a unified text-to-text
  transformer.
\newblock \emph{Journal of Machine Learning Research}, 21:1--67.

\bibitem[{Rajpurkar et~al.(2016)Rajpurkar, Zhang, Lopyrev, and
  Liang}]{rajpurkar2016squad}
Pranav Rajpurkar, Jian Zhang, Konstantin Lopyrev, and Percy Liang. 2016.
\newblock Squad: 100, 000+ questions for machine comprehension of text.
\newblock In \emph{EMNLP}.

\bibitem[{Ravfogel et~al.(2020)Ravfogel, Elazar, Gonen, Twiton, and
  Goldberg}]{ravfogel2020null}
Shauli Ravfogel, Yanai Elazar, Hila Gonen, Michael Twiton, and Yoav Goldberg.
  2020.
\newblock Null it out: Guarding protected attributes by iterative nullspace
  projection.
\newblock In \emph{Proceedings of the 58th Annual Meeting of the Association
  for Computational Linguistics}, pages 7237--7256.

\bibitem[{Schick et~al.(2021)Schick, Udupa, and Sch{\"u}tze}]{schick2021self}
Timo Schick, Sahana Udupa, and Hinrich Sch{\"u}tze. 2021.
\newblock Self-diagnosis and self-debiasing: A proposal for reducing
  corpus-based bias in nlp.
\newblock \emph{arXiv preprint arXiv:2103.00453}.

\bibitem[{Socher et~al.(2013)Socher, Perelygin, Wu, Chuang, Manning, Ng, and
  Potts}]{socher2013recursive}
Richard Socher, Alex Perelygin, Jean Wu, Jason Chuang, Christopher~D Manning,
  Andrew~Y Ng, and Christopher Potts. 2013.
\newblock Recursive deep models for semantic compositionality over a sentiment
  treebank.
\newblock In \emph{Proceedings of the 2013 conference on empirical methods in
  natural language processing}, pages 1631--1642.

\bibitem[{Wang et~al.(2018)Wang, Singh, Michael, Hill, Levy, and
  Bowman}]{wang2018glue}
Alex Wang, Amanpreet Singh, Julian Michael, Felix Hill, Omer Levy, and Samuel
  Bowman. 2018.
\newblock Glue: A multi-task benchmark and analysis platform for natural
  language understanding.
\newblock In \emph{Proceedings of the 2018 EMNLP Workshop BlackboxNLP:
  Analyzing and Interpreting Neural Networks for NLP}, pages 353--355.

\bibitem[{Wang et~al.(2019)Wang, Singh, Michael, Hill, Levy, and
  Bowman}]{wang2019glue}
Alex Wang, Amanpreet Singh, Julian Michael, Felix Hill, Omer Levy, and Samuel~R
  Bowman. 2019.
\newblock Glue: A multi-task benchmark and analysis platform for natural
  language understanding.
\newblock In \emph{7th International Conference on Learning Representations,
  ICLR 2019}.

\bibitem[{Warstadt et~al.(2019)Warstadt, Singh, and
  Bowman}]{warstadt2019neural}
Alex Warstadt, Amanpreet Singh, and Samuel~R Bowman. 2019.
\newblock Neural network acceptability judgments.
\newblock \emph{Transactions of the Association for Computational Linguistics},
  7:625--641.

\bibitem[{Webster et~al.(2020)Webster, Wang, Tenney, Beutel, Pitler, Pavlick,
  Chen, Chi, and Petrov}]{webster2020measuring}
Kellie Webster, Xuezhi Wang, Ian Tenney, Alex Beutel, Emily Pitler, Ellie
  Pavlick, Jilin Chen, Ed~Chi, and Slav Petrov. 2020.
\newblock Measuring and reducing gendered correlations in pre-trained models.
\newblock \emph{arXiv preprint arXiv:2010.06032}.

\bibitem[{Wolf et~al.(2020)Wolf, Debut, Sanh, Chaumond, Delangue, Moi, Cistac,
  Rault, Louf, Funtowicz, Davison, Shleifer, von Platen, Ma, Jernite, Plu, Xu,
  Le~Scao, Gugger, Drame, Lhoest, and Rush}]{wolf-etal-2020-transformers}
Thomas Wolf, Lysandre Debut, Victor Sanh, Julien Chaumond, Clement Delangue,
  Anthony Moi, Pierric Cistac, Tim Rault, Remi Louf, Morgan Funtowicz, Joe
  Davison, Sam Shleifer, Patrick von Platen, Clara Ma, Yacine Jernite, Julien
  Plu, Canwen Xu, Teven Le~Scao, Sylvain Gugger, Mariama Drame, Quentin Lhoest,
  and Alexander Rush. 2020.
\newblock \href {https://doi.org/10.18653/v1/2020.emnlp-demos.6} {Transformers:
  State-of-the-art natural language processing}.
\newblock In \emph{Proceedings of the 2020 Conference on Empirical Methods in
  Natural Language Processing: System Demonstrations}, pages 38--45, Online.
  Association for Computational Linguistics.

\bibitem[{Zhao et~al.(2018)Zhao, Zhou, Li, Wang, and Chang}]{zhao2018learning}
Jieyu Zhao, Yichao Zhou, Zeyu Li, Wei Wang, and Kai-Wei Chang. 2018.
\newblock Learning gender-neutral word embeddings.
\newblock In \emph{EMNLP}.

\bibitem[{Zmigrod et~al.(2019)Zmigrod, Mielke, Wallach, and
  Cotterell}]{zmigrod2019counterfactual}
Ran Zmigrod, Sabrina~J Mielke, Hanna Wallach, and Ryan Cotterell. 2019.
\newblock Counterfactual data augmentation for mitigating gender stereotypes in
  languages with rich morphology.
\newblock In \emph{Proceedings of the 57th Annual Meeting of the Association
  for Computational Linguistics}, pages 1651--1661.

\end{thebibliography}
\bibliographystyle{acl_natbib}

\appendix 

\section{Appendix}
\label{sec:A}

\subsection{Baselines}
\label{sec:A2}
For comparison purposes, we chose two state-of-the-art baselines which focus on debiasing sentence-level pre-trained text encoders in PLMs.

\subsubsection{SENT-DEBIAS}
SENT-DEBIAS \cite{liang2020towards} is an extension of the HARD-DEBIAS method \cite{bolukbasi2016man} to debias sentences for both binary and multi-class bias attributes spanning gender and religion. The key advantage of Sent-D is the contextualization
step in which bias-attribute words are converted into bias-attribute sentences by using a diverse set of sentence templates from text corpora.
Sent-D is a four-step process that involves: identifying words that exhibit biased attributes, contextualizing them in sentences that contain these biases, creating sentence representations, estimating the subspace of the bias represented in the sentences, and debiasing general sentences by removing the projection onto this subspace.
\subsubsection{FairFil}
FairF \cite{cheng2021fairfil} is the first neural debiasing method for pretrained sentence encoders. For
a given pretrained encoder, FairF learns a fair filter (FairFil) network, whose inputs are the
original embedding of the encoder, and outputs are the debiased embedding. Inspired by the
multi-view contrastive learning \cite{chen2020simple}, for each training sentence, FairF first generates an augmentation that has the same semantic meaning but in a different potential bias direction. FairFil is contrastively trained  by maximizing the mutual information between the debiased embeddings of the original sentences and corresponding augmentations. To further eliminate bias from sensitive words in sentences, FairF uses debiasing regularizer, which minimizes the mutual information between debiased embeddings and the sensitive words’ embeddings. 

\subsection{Bias Evaluation Metric}
\label{sec:A3}

Following the prior studies (Sent-D and FairF), we use Sentence Encoder Association Test (SEAT) \citep{may2019measuring} to measure the gender bias scores in the pre-trained language models that trained using Gender-tuning. SEAT extended the Word Embedding Association Test (WEAT; \cite{caliskan2017semantics}) to sentence-level representations. WEAT compares the distance of two sets. Two sets of target words (e.g., \{\emph {family,  child,  parent,...}\} and \{\emph{work, office, profession,...}\} ) that characterize particular concepts $family$ and $career$ respectively.  Two sets of attribute words (e.g., \{\emph{man, he, him,...}\} and \{\emph{woman, she, her,...}\} ) that characterize a type of bias. WEAT evaluates whether the representations for words from one particular attribute word set tend to be more closely associated with the representations for words from one particular target word set. For instance, if the $female$ attribute words listed above tend to be more closely associated with the $family$ target words, this may indicate bias within the word representations.  

Let's denote $A$ and $B$ as sets of attribute words and $X$ and $Y$ the set of target words. As described in \citep{caliskan2017semantics} the WEAT test statistic is:

\begin{equation}
\label{eq:2}
s (X, Y, A, B) = \sum_{x \in X} s(x, A, B) - \sum_{y \in Y} s(y, A, B)
\end{equation}
where for a specific word $w$ , $s(w, A, B)$ is defined as the difference between $w$'s mean cosine similarity with the words from $A$ and $w$'s mean cosine similarity with the word from $B$. They report an effective size given by:

\begin{equation}
\label{eq:3}
d = \frac{\mu([s(x, A, B)]_{x \in X} - \mu([s(y, A, B)]_{y \in Y})}{\sigma{([s(t, X, Y)]_{t \in A\cup B}})}
\end{equation}
where $\mu$ and $\sigma$ denote the mean and standard deviation respectively. Hence, an effect size closer to zero represents smaller degree of bias in the word representation. The SEAT test extended WEAT by replacing the word with a collection of  template sentences (i.e., \emph{"this is a [word]", "that is a [word]"}). Then the WEAT test statistic can be computed on a given sets of sentences including attribute and target words using sentence representations from a language model.   

\subsection{PLMs}
\label{sec:A4}
Two widely used  pre-trained language models have been chosen for this study, BERT-base \citep{devlin2019bert}and RoBERTa-base \citep{liu2019roberta}.  BERT-base is a bidirectional encoder with 12 layers and 110M parameters that is pre-trained on 16GB of text. RoBERTa-base has almost the same architecture as BERT but is pre-trained on ten times more data (160GB) with significantly more pre-training steps than BERT.

\subsection{Datasets}
\label{sec:A5}
We conducted empirical studies on the following three tasks from the GLUE benchmark \cite{wang2019glue}: \\
(1) \textbf{SST-2}: Stanford Sentiment Treebank is used for binary classification for sentences extracted from movie reviews \citep{socher2013recursive}. It contains 67K training sentences. \\
(2) \textbf{CoLA}: Corpus of Linguistic Acceptability \citep{warstadt2019neural}  consists of English acceptability judgment. CoLA contains almost 9K training examples.\\
(3) \textbf{QNLI}: Question Natural Language Inference \citep{wang2018glue} is a QA dataset which is derived from the Stanford Question Answering Dataset \citep{rajpurkar2016squad} and used for binary classification. QNLI contains 108K training pairs.  \\
Also, we use the feminine and masculine word lists created by \citep{zhao2018learning} for gender-word perturbation in Gender-tuning.

\subsection{Hyperparameters}
\label{sec:A6}
The hyperparameters of the models, except batch size, are set to their default (https://github.com/huggingface/transformers) values (e.g., epoch $=$ 3, learning-rate $=$ $2\times10^{-5}$, and etc.). After trying several trials run, the batch size has been selected among $\{8, 16, 32\}$. 
We empirically selected the optimal value for $\alpha$ by a grid search in $0$ < $\alpha$ < $1$ with 0.1 increments. For each downstream task, the best value of $\alpha$ sets to 0.7. 
All experiments were performed with three training epochs and using an NVIDIA V100 GPU. 
\subsection{Related Works}
\textbf{ Debiasing Database}; The most straightforward approach for reducing the social biases in the training corpora is bias-neutralization. In this way, the training corpus is directly re-balanced by swapping or removing bias-related words and counterfactual data augmentation (CDA) \citep{zmigrod2019counterfactual,dinan2020queens,webster2020measuring,dev2020measuring, barikeri2021redditbias}. Also, \citet{gehman2020realtoxicityprompts} proposed domain-adaptive pre-training on unbiased corpora. Although the results showed these proposed methods mitigated the social biases in the pre-trained models,   they need to be re-trained  on a larger scale of the corpora. For example, \citet{webster2020measuring} proposed a CDA that needs an additional 100k steps of training on the augmented dataset. Data augmentation and collecting a large-scale unbiased corpus are both computationally costly.\\ \\
\textbf{Debiasing Embedding}; There are several solutions for debiasing static word embedding   \citep{bolukbasi2016man, kaneko2019gender,manzini2019black, ravfogel2020null} and debiasing contextualized word-embedding \citep{caliskan2017semantics, brunet2019understanding} and sentence-embedding \citep{liang2020towards, cheng2021fairfil}. Compared to debiasing static word embedding, where the semantic representation of a word is limited to a single vector, contextualized word/sentence embedding models are more challenging \citep{kaneko2019gender}. 
Since the key to the pre-trained language models' success is due to powerful embedding layers \citep{liang2020towards}, debiasing embedding might affect transferring of the accurate information and performance of these models on the downstream tasks. Also, they need some pre-training for debiasing the embedding layer before fine-tuning on downstream tasks.

\end{document}